\documentclass{sig-alternate}

\graphicspath{{figures/}}

\usepackage{tabularx}

\newcommand{\AoS}{ATNoSFERES}

\begin{document}
%
\conferenceinfo{GECCO'05,} {June 25--29, 2005, Washington, DC, USA.}
\CopyrightYear{2005}
\crdata{1-59593-010-8/05/0006}

\title{ATNoSFERES revisited}
\numberofauthors{3}

\author{
\alignauthor Samuel Landau\\
       \affaddr{Équipe TAO - INRIA Futurs}\\
       \affaddr{LRI - Bât. 490}\\
       \affaddr{Université de Paris-Sud}\\
       \affaddr{91405 Orsay, France}\\
       \email{Samuel.Landau@lri.fr}
\alignauthor Olivier Sigaud\\
       \affaddr{LIP6}\\
       \affaddr{Université Pierre et Marie Curie}\\
       \affaddr{Paris, France}\\
       \email{Olivier.Sigaud@lip6.fr}
\alignauthor Marc Schoenauer\\
       \affaddr{Équipe TAO - INRIA Futurs}\\
       \affaddr{LRI - Bât. 490}\\
       \affaddr{Université de Paris-Sud}\\
       \affaddr{91405 Orsay, France}\\
       \email{Marc.Schoenauer@lri.fr}
}

\date{}
\maketitle

\begin{abstract}
\AoS\ is a Pittsburgh style Learning Classifier System (LCS) in which the rules are
represented as edges of an Augmented Transition Network.  Genotypes are
strings of tokens of a stack-based language, whose execution  builds
the labeled graph.
The original ATNoSFERES, using a bitstring to represent the language
tokens, has been favorably compared in previous work to 
several Michigan style LCSs
architectures in the context of Non Markov problems.
Several modifications of \AoS\ are proposed here: the most
important one conceptually being a representational change: each token is now
represented by an integer, hence the genotype is a string of
integers; several other modifications of the underlying grammar
language are also proposed.
The resulting ATNoSFERES-II is validated on several standard animat
Non Markov problems, on which it outperforms all previously published
results in the LCS literature. The reasons for these improvement are
carefully analyzed, and some assumptions are proposed on the underlying
mechanisms in order to explain these good results.
\end{abstract}




\section{Introduction}

The Pittsburgh versus Michigan debate has been present in the Learning
Classifier Systems (LCSs) community since the very beginning of this
research area (see \cite{wilson89} for a presentation). On the one hand, most
recent works in the LCS domain have chosen the Michigan approach \cite{ECJ:SpecialIssue2003}.
The \AoS\ architecture proposed in
\cite{Landau:GECCO:2002,landau:IWLCS:2003}, on the other hand, is a 
Pittsburgh style architecture dedicated to the resolution of non
Markov problems. It has been compared to several LCSs, in
particular in \cite{Landau-Sigaud:IS:2004}, where the authors
concluded a systematic comparison by the claim that the \AoS\ 
approach was more robust and more general than 
its Michigan style opponents, but that it was also several orders of
magnitude slower.

This paper investigates the effect of several modifications to the
original \AoS\ approach on its  global efficiency. Instead of
being represented by a bitstring, the genotype, that encodes a series
of tokens of the chosen stack-base graph-building language, is now a
string of integers. This new representation greatly impacts on the
mutation, that can now be naturally made uniform on the set of
tokens. Moreover, all \emph{stack} tokens of the underlying language
can now operate  on specific
data on the stack. Finally, the random default behavior when no edge
is eligible in the current node of the
automaton is suppressed, and the contradictions in the conditions of all edges
of the control graph are filtered out.
Because all those changes tend to make \AoS\ representation and
behavior closer to the actual semantic of the underlying search
space -- that of the ATNs -- 
it is hoped that the resulting algorithm, termed ATNoSFERES-II,
achieves at least as good results as ATNoSFERES, but much faster.

The paper is organized as follows: the next section gives a
short overview of the previously published versions of ATNoSFERES.
Section~\ref{sec:modifications}  details the modifications
introduced in this paper. Section \ref{sec:experiments} 
indicates the
experimental conditions used to test ATNoSFERES-II on
classical Non Markov animat problems.
Section \ref{sec:results} presents the results obtained by
ATNoSFERES-II, compares them to the state-of-the-art LCS algorithms
and produces statistical significance tests of the
improvements. Finally, all those results are discussed in
section~\ref{sec:discussion}, and some conclusions are draw on the
impact of the proposed modifications.

\section{Overview of \AoS
\label{sec:overview}}

\subsection{The bitstring representation}
\label{sub:approach}

The \AoS\ model~\cite{landau:picault:atnosferes}  is designed to generate the control
architecture of agents thanks to an evolutionary algorithm.
The control architecture itself is represented as an
ATN~\footnote{Augmented Transition Network \cite{woods:atn}} graph where 
nodes represent states and edges represent transitions of an automaton.
The graph describing the behaviors is built by interpreting a genotype
as a program in a stack-based language~\cite{landau:picault:SBGE:eng} that 
proceeds by adding nodes and edges to a basic structure initially containing only 
the \textit{Start} and \textit{End} nodes. 
This genotype is a bitstring in the original \AoS\  model.

\begin{table*}
\begin{center}
\small
\begin{tabular}{|*{4}{l l|}}
\hline
000000 & \textit{swap all|node} & 000001 & \textit{swap all|node } & 000010 &
\textit{swap all|label} & 000011 & \textit{swap all|label}\\
000100 & \textit{dup label} & 000101 & \textit{dup label} & 000110 &
\textit{dup node} & 000111 & \textit{dup node}\\
001000 & \textit{del label} & 001001 & \textit{del label} & 001010 &
\textit{del node} & 001011 & \textit{del node}\\
001100 & \textit{roll all|node} & 001101 & \textit{roll all|label} & 001110 &
\textit{unroll all|node} & 001111 & \textit{unroll all|label}\\
010000 & \textit{node} & 010001 & \textit{node} & 010010 &
\textit{node} & 010011 & \textit{node} \\
010100 & \textit{connect} & 010101 & \textit{connect} & 010110 &
\textit{connect} & 010111 & \textit{connect} \\
011000 & \textit{connect self} & 011001 & \textit{connect self} & 011010 &
\textit{connect self} & 011011 & \textit{connect start} \\
011100 & \textit{connect start} & 011101 & \textit{connect start} &
011110 & \textit{connect end} & 011111 & \textit{connect end}\\
100000 & \textit{goN!} & 100001 & \textit{goS!} &
100010 & \textit{goW!} & 100011 & \textit{goE!}\\
100100 & \textit{goNE!} & 100101 & \textit{goSE!} &
100110 & \textit{goNW!} & 100111 & \textit{goSW!}\\
101000 & \textit{emptyN?} & 101001 & \textit{foodN?} &
101010 & \textit{treeN?} & 101011 & \textit{emptyS?}\\
101100 & \textit{foodS?} & 101101 & \textit{treeS?} &
101110 & \textit{emptyW?} & 101111 & \textit{foodW?}\\
110000 & \textit{treeW?} & 110001 & \textit{emptyE?} &
110010 & \textit{foodE?} & 110011 & \textit{treeE?}\\
110100 & \textit{emptyNE?} & 110101 & \textit{foodNE?} &
110110 & \textit{treeNE?} & 110111 & \textit{emptySE?}\\
111000 & \textit{foodSE?} & 111001 & \textit{treeSE?} &
111010 & \textit{emptyNW?} & 111011 & \textit{foodNW?}\\
111100 & \textit{treeNW?} & 111101 & \textit{emptySW?} &
111110 & \textit{foodSW?} & 111111 & \textit{treeSW?}\\
\hline
\end{tabular}
\end{center}
\caption{\footnotesize The genetic code used in the experiments, for 6-bit
codons. 
This is the mapping from bitstrings/integers (in binary representation) to tokens.
 Some codes for the swap,
roll and unroll stack tokens have two alternative behaviors, 
either \emph{all} or \emph{node}/\emph{label} (see the text, section
\ref{sub:All_vs_NodeLabel}). 
Also note that, due to the ATNoSFERES requirement that there are $2^N$
tokens for some $N$, 
some tokens are represented more than once (e.g. \emph{swap all}
is represented 4 times), inducing a slight bias toward those nodes.}
\label{tab:genetic-code}
\vspace{-10pt}
\end{table*}

The graph-building process operates in two steps.
First, during {\em translation}, the genotype is translated into a sequence
of tokens (see table~\ref{tab:genetic-code}).
Second, during {\em interpretation}, the token interpreter is fed with the token stream produced by the translator.
The tokens are interpreted one by one as instructions of a 
robust programming language dedicated to graph building.
The interpretation of each successive token operates on a stack in
which parts of the future graph are stored.
The construction of the graph takes place during this interpretation
process, by creating nodes and connections between nodes.
When all tokens have been interpreted, the nodes (each one carrying
connections to other nodes) are popped from the stack and the graph is
ready to use.

\begin{figure}[ht]
\center \includegraphics[width=.9\linewidth,angle=0]{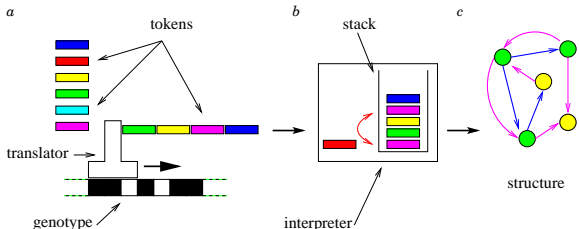}
\vspace{-12pt}
\caption{Principles of the genetic expression used to produce
the behavioral graph from the bitstring genotype. The string is
first decoded into tokens (\textit{a}), which are interpreted in 
a second step as instructions (\textit{b}) to create nodes, edges, and 
labels. Finally, when all tokens have been interpreted, the
unused conditions and actions remaining in the stack are added
into the structure and the structure is popped from the stack (\textit{c}).}
\label{fig:stack}
\end{figure}

There are three kinds of tokens: \emph{action and condition} tokens,
such as \emph{SW!} or \emph{foodNE?}, that are
action and condition labels for the edges of the ATNs, specific to the
actions (movements) and perceptions (what is in a nearby cell) of the
agent in the maze environments; \emph{graph structure} tokens, such
as \emph{node} or \emph{connect}, that are instructions to push nodes
and connect them with edges, using the labels in the stack; and
finally \emph{stack} tokens, such as \emph{swap all} or \emph{roll
node}, that manipulate the stack either as a whole (the \emph{all}
family of stack tokens) or by acting only on one of the two kinds of data
present in the stack (the \emph{node} and the \emph{label} families of
stack tokens).

\subsection{The evolutionary framework}
\label{sub:evolution}
A generation of ATNoSFERES 
classically starts with the selection of the parents that will be
selected for reproduction, then applies variation operators (crossover and mutation)
to these lucky parents.

The selection is a rank-based truncated selection that proceeds as
follows: First, the $n$ best individuals of the population (of size
$P$) are retained (deterministic truncation), and sorted by descending
fitness. They will be used to generate $P-n$ offspring that will
complete the population. Then, in order to generate each pair of offspring, 
two  parents are selected from those $n$ best, based on
a rank-based exponential selection: if the best parent is associated
with a bias $b$, the $i^{th}$ best parent is associated with bias $b.c^i$,
for some $c\in ]0,1[$. 
Thus, of course $b$, $c$ and $n$ must satisfy $\sum_{i=0}^{n-1}{b.c^i}=1$.
In all experiments described throughout this paper, $P=300$, $n=60$,
and $c$ was chosen so that the selection pressure after truncation
(bias from the best to the worst of the $n$ parents) is 2 
(i.e.~$c=\sqrt[60]{0.5}$). 

After the selection described above, 
the genotype of two offspring is produced by a
2-points crossover between the two genotypes, where
crossover points must be at the border between two tokens.

Additionally, in the original \AoS\ approach, 
two different mutation strategies were used: 
classical bit-flip mutation, and random insertions
or deletions of one codon (i.e.~one token of the language at
hand). However, using add/delete mutations did not improve the performances
of ATNoSFERES, and only the bit-flip mutation was ever used.
This modifies 
the sequence of tokens produced by translation,
so that the complexity of the graph itself may change.
Nodes or edges can hence  be added or removed by the 
evolutionary process, as can condition/action labels
on the edges.

\subsection{The evaluation function}
\label{sub:evaluation}
The fitness of each genotype is assessed by first building the ATN, as
described above, then by putting this ATN as the controller of an
agent and evaluating the behavior of the agent in a (Non Markov)
environment. There are three sources of
non-determinism in the use of the ATN, therefore,
each parent is re-evaluated together with the offspring, and
the fitness is averaged over successive evaluations (by calculating
the mean fitness over all evaluations).

To evaluate an agent, the ATN graphs are used as follows for at most a
fixed number of time steps:
\begin{itemize}
\item At the beginning (when the agent is initialized),
the agent is at the \textit{Start} node (S).
\item At each time step, the agent crosses an edge:
        \begin{enumerate}
        \item It computes the set of eligible edges among 
        those starting from the current node. An edge is eligible when 
        either it has no condition label 
        or all the conditions on its label are simultaneously true. 
        \item An edge is chosen in this set. 
The first versions of
\AoS\/ were selecting one edge randomly (this is the first
source of non-determinism), but it was
found that a deterministic  choice (e.g.~choose the first edge
in the list of eligible edges) held better results. 
        If the set of eligible edges is
        empty, then an action is chosen randomly over all possible
        actions, and the current node remains unchanged (this is the
	second source of non-determi\-nism).
        \item The actions on the label of the elected edge are
        sequentially performed by the system. Assuming that only one
        action can be 
        performed at a time, only the last action is actually performed.
        When the action part of the label is empty, an action is
        chosen randomly. This is the third and last source of non-determinism.
However, in all experiments described in this paper, the action part
of all edges 
will contain at most one action, in order to simplify the comparison
with LCSs. 
        \item The target node of the elected edge becomes the new
          current node. 
        \end{enumerate}
\item The agent stops when it reaches the \textit{End} node (E). 
This node is a general feature of the model and may never be reached,
either if the agent loops infinitely or if the experiment is stopped
before.
\end{itemize}

\section{ATNoSFERES-II}\label{sec:modifications}

This section introduces the modifications that have been brought to the
original ATNoSFERES model described in the previous section. In order
to be able to assess the effects of each modification independently of
one another, all combinations will be made possible, even if the
algorithm referred to as ATNoSFERES-II uses them all.

\subsection{Integer representation}
\label{sub:BitpFlip_vs_Uniform}
Instead of coding a genome as a bitstring, a string of integers is now
used, where each integer is an index encoding a token (in the range
$[0,\#tokens]$). There might be repeated tokens, which is the
case in our experiments.
First, this simplifies the translation process (the 
decoding of a bitstring into an integer disappears). 
However, and because crossover was only allowed at the boundary
between tokens in the bitstring genotype of the original \AoS\/, the
only visible effect of this deep modification is at the mutation
level: it is now possible to use the uniform mutation operator
that replaces a given token by another one uniformly, i.e. with equal
chance for all tokens in the list.

Another possibility could be to define a distance among 
the tokens based on their semantics (their effect on the structure
being built), in order to bias the mutation. Such a
bias would be used to smooth mutation strength by having smaller
effects on the phenotype change. However, it is clear that the bitflip
mutation used in the 
original \AoS\ was in fact equivalent 
to such a weighted mutation, but the mutation biases were dependent on the
order of the tokens in the token list (e.g.~by having condition and
action tokens at the second half of the list, the first bit was a flag for
condition/action tokens). The induced mutation biases were thus
completely arbitrary with respect to the problem at hand.
Another important difference is that 
the total number of encoded tokens no longer has to be $2^N$ for some
$N$ (where $N$ is the number of bits used to represent a token).

Note that a similar effect could also have been obtained
in the bitstring context by increasing the bitstring length
and decoding each token using a {\em modulo} function, as is
theoretically proved in \cite{EA03Nicolau}. 

In the following, and because the only visible effect of this change
of representation is the actual change of mutation, the original \AoS\
approach will be referred to as \emph{BitFlip} while the ATNoSFERES-II
approach will be termed \emph{Uniform}.

\subsection{Default node action}
\label{sub:Random_vs_Finish}
As stated in section~\ref{sub:evolution}, in the original version of
ATNoSFERES, when no edge could be elected, an action was chosen
randomly -- this was referred to as the second source of
non-determinism. However, this results in non-deterministic
performance of the algorithm.

In ATNoSFERES-II, a drastic strategy is used in such a case: when no
edge is eligible, the evaluation of the agent stops and {\em FAIL} is
returned. As a consequence, the
remaining number of time steps for the current test is decreased to
zero, and the fitness for this test is null
(the overall fitness is the sum over several tests, see section
\ref{sec:experiments}). 
In the following, this choice of abruptly ending the test when no edge
is eligible will 
be referred to as \emph{Finish}, while the random choice of ATNoSFERES
will be called \emph{Random}.

We must emphasize that the third source of non-determi\-nism (performing a
random action when crossing an edge with no action label) seems
experimentally to have much less impact on non-deterministic
performances. We observed that the few such edges still present in the
population after some generations were never eligible.

\subsection{Contradiction filtering}
\label{sub:Contradiction_vs_NoContradiction}
In the original version of ATNoSFERES, an edge of the control
graph could eventually be labeled with contradictory conditions,
resulting in its permanent ineligibility.
Such a situation cannot happen in LCSs, since the 
condition part of a classifier cannot contain contradictions.
In the present work,  the occurrence of 
contradictory conditions in the label of edges is prevented by 
forbidding more than one condition 
specifying the value of the same attribute in the condition: once an
attribute has been used in a given label, all subsequent tokens
involving that attribute are ignored for that label.
The use of this mechanism will be 
referred to as \emph{No-Contradiction}, in contrast with the
\emph{Contradiction} original ATNoSFERES algorithm.

\subsection{Behavior of swap/roll/unroll tokens}
\label{sub:All_vs_NodeLabel}

In the original version of ATNoSFERES, the swap, roll and unroll stack
tokens 
could operate without discrimination on any data in the stack, whereas
some other stack tokens could only operate either on the
nodes or on the labels: e.g. \emph{dup node} only
operated on nodes, and  \emph{dup label} only on labels. 
This ``typing'' of tokens is extended in ATNoSFERES-II to all
available stack tokens, in order to facilitate structural
manipulations of the automata being built.
The previous set of
the untyped swap/roll/unroll tokens will be referred to as \emph{all}, while the
typed ones will be referred to as \emph{node/label}.

\section{Experimental settings}
\label{sec:experiments}

\subsection{Representation}
Our experimental set-up is the same as in
\cite{Landau-Sigaud:IS:2004}, described in section~\ref{sub:approach},
except for the new features of ATNoSFERES-II. We used a 1\% bias for
mutation probabilities.
In particular, since one of the goals of the present 
experiments is 
to compare the uniform and the bit-flip mutations on a fair basis, the
distribution of the set of tokens is also the same (see
table~\ref{tab:genetic-code}), except that, of course, integer- and
not bit-strings 
are used in the \emph{Uniform} approach of ATNoSFERES-II (section
\ref{sub:BitpFlip_vs_Uniform}).
The other difference, as pointed out in section
\ref{sub:All_vs_NodeLabel}, lies in the use of ``typed'' \emph{swap},
\emph{roll} and \emph{unroll} tokens (see alternatives in
table~\ref{tab:genetic-code}), when it comes to compare \emph{Node/Label} against
\emph{All}.

Capitalizing on previous published experiments, we do not start
with as many genome lengths as before
\cite{landau:IWLCS:2003,Landau-Sigaud:IS:2004}. Some
preliminary tests with lengths of 250, 300 and 350 tokens on a subset
of parameter combinations demonstrated that, in this range, the
genome lengths had no statistically significant impact
on the final performance. 
From thereon, and with an abusive generalization to all environments,
we shall consider
that the average results obtained with this limited range of lengths
can be extended to other lengths for all environments.

Also, in the next section, the graphical representation of the graphs
is different from that of \cite{landau:IWLCS:2003,Landau-Sigaud:IS:2004},
in order to make the figures more
readable (e.g. see figure~\ref{fig:maze10_best_ATN}). First, the condition part is presented above the action
part. Then, each
condition is here prefixed by a letter representing the perception of
the agent: \textbf{f}ood, \textbf{t}ree, or \textbf{e}mpty. 
The \emph{Start} node is
labeled \emph{0}, and the \emph{End} node is the node labeled with the
largest number.
If no condition is present on an edge, any perception will match.
Nodes and edges are represented as usual by circles and directed arrows.

\subsection{The environments and the fitness}

Three different environments, respectively {\em Maze10} \cite{lanzi98analysis},
\emph{E1} and {\em E2} \cite{metivier:lattaud:ACS},
 have been used to validate and test the
modifications of ATNoSFERES proposed in section
\ref{sec:modifications}. 
For all these environments, the ``optimal policy'' is the
best policy that can be found with the standard limited perception
used in the so called ``woods'' experiments \cite{wilson:animat:1}
without any limit on the number of memory bits.
The number of steps to find the food from each cell given by this
optimal policy has been presented in \cite{Landau-Sigaud:IS:2004}.

For each given environment, each automaton is started once in each
cell, and its fitness is measured as the average number of steps
it takes to find the food, averaged over all cells (to be minimized).

\section{Results}
\label{sec:results}

Because we did not want to make any assumption about the
usefulness of each 
on the modifications to the original ATNoSFERES proposed in section 
\ref{sec:modifications}, 
systematic experiments were conducted in 
each of the three environments in order to evaluate the 16 possible 
combinations of each of the four modifications, with 50 independent
runs per setting. 

\subsection{Best solutions found}

This section will present the best overall results obtained in each
environment.
In each case, both the automaton and a
picture of its performance 
with respect to the optimal policy is given.
All those automata have been obtained using the \emph{Uniform} and
\emph{No contradiction} modifications -- the other two will be
detailed in each subsection.

\subsubsection{Maze10}

\begin{figure}[htbp]
\begin{center}
\includegraphics[width=.9\linewidth,angle=0]{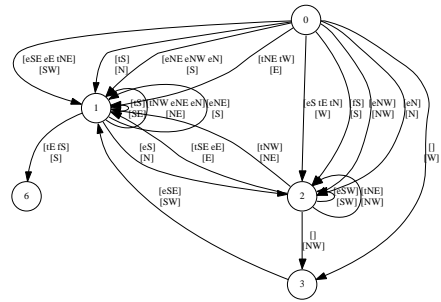}
\vspace{-12pt}
\end{center}
\caption{Best automaton found for the Maze10 environment ($5.11$
steps to food; an optimal policy requires $5.05$). 19 edges
(3 nodes) that are never elected (reached) are not
represented, for the sake of readability. \label{fig:maze10_best_ATN}}
\end{figure}

\begin{figure}[htbp]
\begin{center}
\includegraphics[width=.4\linewidth,angle=0]{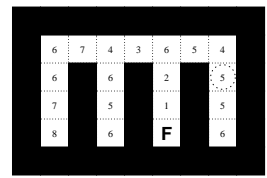}
\vspace{-12pt}
\end{center}
\caption{Performance of the best policy found for the Maze10
  environment (by the automaton of figure \ref{fig:maze10_best_ATN}). 
It is only one step worse, and on a single cell,
than the optimum (dotted
circle). The rest is optimal. \label{fig:maze10_best_policy}}
\end{figure}

In Maze10, the best result of ATNoSFERES-II
(figure~\ref{fig:maze10_best_ATN}) has a policy that is is only one
step  longer than the optimal policy (see
figure~\ref{fig:maze10_best_policy}), when the {\em Start} cell is 
just below the NE corner of the maze.
It was found using 300 codons, and the {\em Node/Label} and {\em
  Finish}  parameters.
As far as we can tell, no LCS has ever
performed so well on this problem.
The average number of steps to food is $5.11$,
vs. $5.61$ for \AoS\ in \cite{Landau-Sigaud:IS:2004}.

\subsubsection{E1}

\begin{figure}[htbp]
\begin{center}
\includegraphics[width=1.0\linewidth,angle=0]{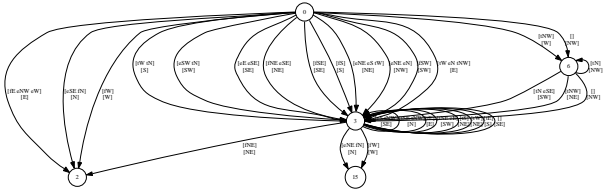}
\vspace{-12pt}
\end{center}
\caption{Best automaton found for the E1 environment ($2.90$
steps to food; an optimal policy requires $2.81$). 30 edges
(11 nodes) that were never elected (reached) are not
represented, for the sake of readability. \label{fig:e1_best_ATN}}
\end{figure}

\begin{figure}[htbp]
\begin{center}
\includegraphics[width=.3\linewidth,angle=0]{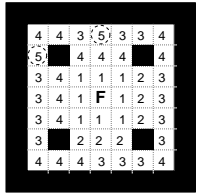}
\vspace{-12pt}
\end{center}
\caption{Best policy found for the E1 environment (by the automaton of figure \ref{fig:e1_best_ATN}). Only two cells are
  only two time steps away from the optimum (dashed
  circles).  The rest is optimal. \label{fig:e1_best_policy}} 
\end{figure}

In environment E1, the best policy found is only 4 steps longer than
the optimal 
one (figure~\ref{fig:e1_best_policy}).
It was found with a chromosome length of 250, and using the {\em All}
and {\em  Finish} settings.
Again, as far as we are aware, no LCS
has ever performed so well on this problem:
The average number of steps to food is $2.90$,
vs. $3.3$ for ACS in \cite{metivier:lattaud:ACS}.

\subsubsection{E2}

\begin{figure}[htbp]
\begin{center}
\includegraphics[width=1.0\linewidth,angle=0]{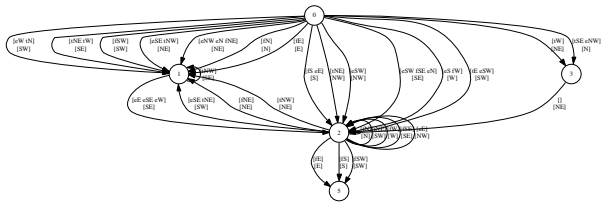}
\vspace{-12pt}
\end{center}
\caption{Best automaton found for the E2 environment ($3.29$
steps to food; an optimal policy requires $2.98$). 13 edges
(1 node) that were never elected (reached) are not
represented, for the sake of readability.\label{fig:e2_best_ATN}}
\end{figure}

\begin{figure}[htbp]
\begin{center}
\includegraphics[width=0.3\linewidth,angle=0]{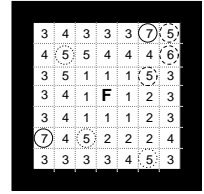}
\vspace{-12pt}
\end{center}
\caption{Best policy found for the E2 environment (by the automaton of
figure \ref{fig:e2_best_ATN}). Three 
cells are one time step away from the optimum (dotted circles),
three cells are two time steps away
from the optimum (dashed circles), and two cells are three time steps
away from the optimum (solid circles). The rest is optimal.\label{fig:e2_best_policy}}
\end{figure}

In environment E2, the best policy found is 15 steps away from the
optimum (figure~\ref{fig:e2_best_policy}). 
Its genotype length is 350, and again the {\em All} and {\em Finish}
strategies were used.
As for the
two other environments, as far as we can tell, no LCS has ever
performed so well on this problem: The average number of steps to food is
$3.29$, vs. $3.58$ for \AoS\ in \cite{Landau-Sigaud:IS:2004}.

\subsection{Validation of the different modifications \label{sub:signif}}

The noticeable improvements of the off-line best results reported in
the previous section when using ATNoSFERES-II have been obtained using
(some of) the 
modifications presented in section~\ref{sec:modifications}. The
present section investigates which of the corresponding modifications
has a significant impact on the overall performance of the
algorithm in average.

\begin{table}[ht]
\begin{center}
\small
\begin{tabular}{r|c|c|c|}
 & Maze10 & E1 & E2 \\
\hline
Uniform & 8.2 & 3.7 & 4.3 \\
BitFlip &  12 &   5 & 6.4 \\
 p-value & $\approx\mathbf{0}$ & $\approx\mathbf{0}$ & $\approx\mathbf{0}$ \\
\hline
All &  10 & 4.4 & 5.4 \\
Node/Label & 9.7 & 4.4 & 5.3 \\
p-value & \bf{0.02} & 1.0 & 0.9 \\
\hline
Contradiction &  10 & 4.4 & 5.4 \\
No contradiction & 9.8 & 4.3 & 5.3 \\
p-value & 0.2 & \bf{0.05} & 0.4 \\
\hline
Random & 9.7 & 4.3 & 5.3 \\
Finish &  10 & 4.4 & 5.4 \\
p-value & \bf{0.04} & 0.6 & 0.2 \\
\hline
\end{tabular}
\caption{\footnotesize Average number of steps to food for each parameter value and environment, and
T-test p-value for each couple of alternative parameter values.
A p-value lower than 0.05 means a statistically significant difference
(in bold).
\label{tab:p-values}} 
\end{center}
\end{table}

\begin{table}[ht]
\begin{center}
\small
\begin{tabular}{r|c|c|c|}
\bf{Uniform} & Maze10 & E1 & E2 \\
\hline
All & 8.6 & 3.7 & 4.3 \\
Node/Label & 7.7 & 3.7 & 4.4 \\
p-value & $\approx\mathbf{0}$ & 0.7 & 0.6 \\
\hline
Contradiction & 8.1 & 3.8 & 4.4 \\
No contradiction & 8.3 & 3.7 & 4.3 \\
p-value & 0.4 & 0.06 & 0.2 \\
\hline
Random & 7.9 & 3.7 & 4.3 \\
Finish & 8.4 & 3.7 & 4.4 \\
p-value & \bf{0.01} & 0.9 & \bf{0.03} \\
\hline
\end{tabular}
\caption{\footnotesize Average mean for each parameter value and environment, and
T-test p-value for each couple of alternative parameter values, assuming Uniform.\label{tab:p-values2}} 
\end{center}
\end{table}

\begin{table}[ht]
\begin{center}
\small
\begin{tabular}{r|c|c|c|}
\bf{Uniform \& Random} & Maze10 & E1 & E2 \\
\hline
All & 8.1 & 3.7 & 4.2 \\
Node/Label & 7.7 & 3.8 & 4.3 \\
p-value & 0.1 & 0.5 & 0.4 \\
\hline
Contradiction &   8 & 3.8 & 4.3 \\
No contradiction & 7.8 & 3.6 & 4.2 \\
p-value & 0.5 & \bf{0.03} & 0.4 \\
\hline
\end{tabular}
\caption{\footnotesize Average mean for each parameter value and environment, and
T-test p-value for each couple of alternative parameter values,
assuming Uniform and Random.\label{tab:p-values3}} 
\end{center}
\end{table}

Table~\ref{tab:p-values} analyzes the results according to each couple
of parameter values, that is, for each value of a given parameter
(e.g. \emph{Uniform} or \emph{BitFlip}), the results are averaged over
the 8 possible values of the other 3 parameters.
It  clearly appears that \emph{Uniform} mutation
produces 
significantly better results than \emph{BitFlip} in all three
environments. The statistical significance is not so clear for the
other strategies,  
except for  \emph{Node/Label} and \emph{Random} for Maze10, and
\emph{No contradiction} for E1. 

Hence in the remaining of this section, \emph{Uniform}
is assumed, and the analysis is carried over on the 8 sets
of experiments for which \emph{Uniform} was used.
Table~\ref{tab:p-values2} shows the results of this analysis, 
and it appears that  \emph{Random} gives significantly better  results
for Maze10 
and E2, but not for E1. Other parameters do not produce
significantly different results, with the exception of
\emph{Node/Label} for Maze10. 

Assuming now \emph{Uniform}
and \emph{Random}, \emph{No contradiction} still gives significantly
better results for the E1 environment (see
table~\ref{tab:p-values3}). 
So even if the \emph{Contradiction}/\emph{No contradiction}
parameters do not produce significantly different results for all
environments,  \emph{No contradiction} nevertheless gives
better average mean results in all experiments, whether considered
alone, or in conjunction with either \emph{Uniform} alone or 
\emph{Uniform} and \emph{Random}. Furthermore,  all the best known 
solutions found for the tested 
environments use the  \emph{No contradiction} value.

\begin{figure}[ht]
\center \includegraphics[width=0.6\linewidth,angle=-90]{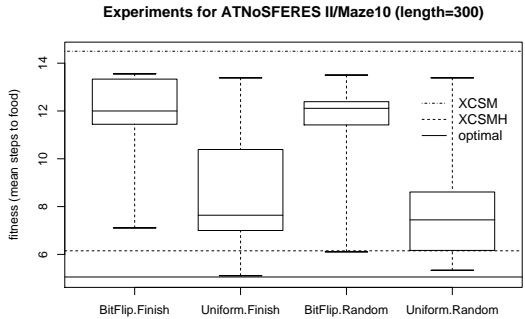}
\vspace{-12pt}
\caption{Results for the Maze10 environment, according to the mutation
and default node action variables. See section~\ref{sub:signif} for
details and analysis.\label{fig:maze10_results}}
\end{figure}

\begin{figure}[ht]
\center \includegraphics[width=0.6\linewidth,angle=-90]{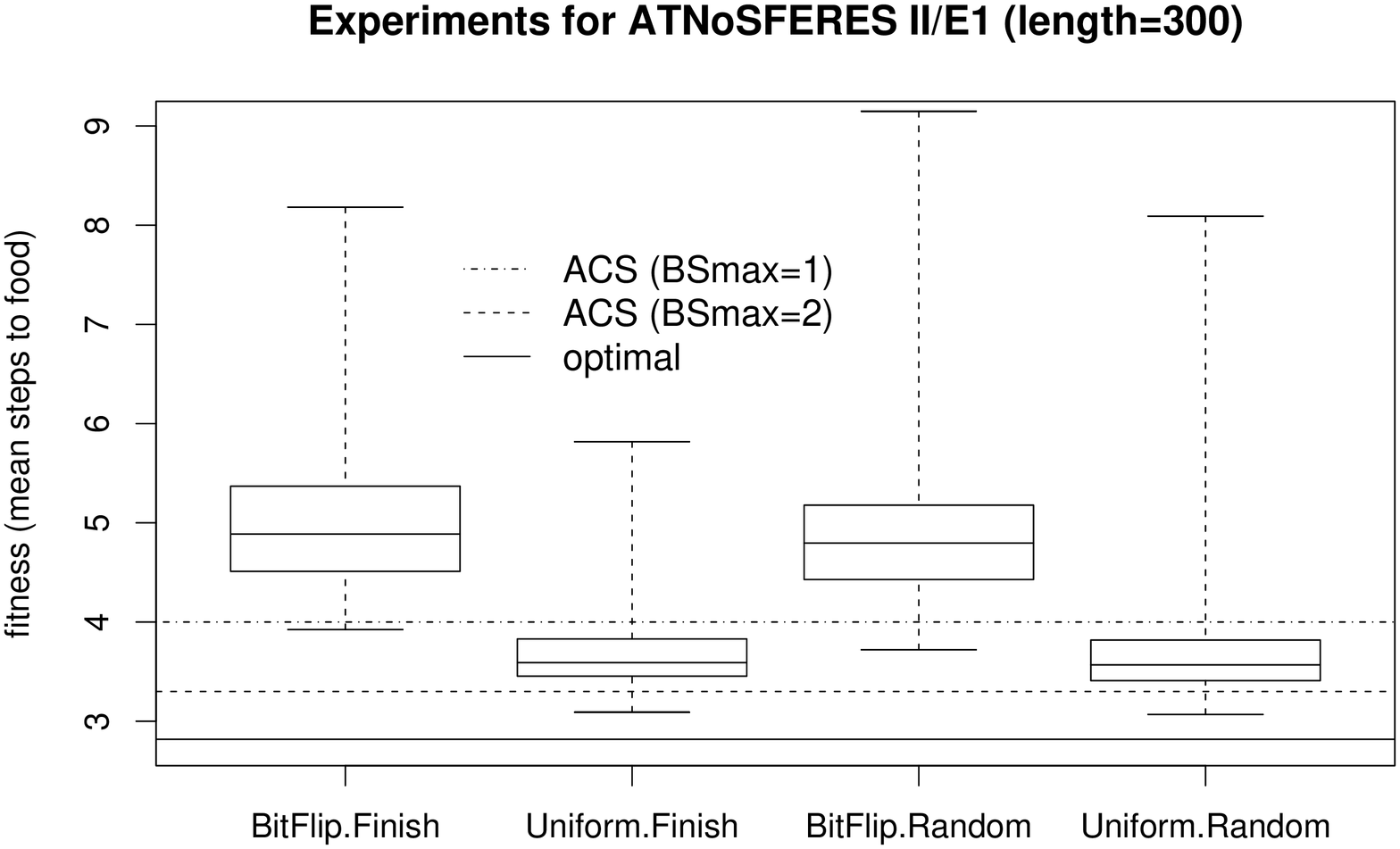}
\vspace{-12pt}
\caption{Results for the E1 environment, according to the mutation and
default node action variables. See section~\ref{sub:signif} for
details and analysis.
 \label{fig:e1_results}}
\end{figure}

\begin{figure}[ht]
\center \includegraphics[width=0.6\linewidth,angle=-90]{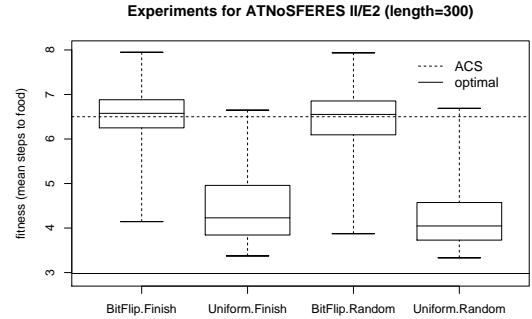}
\vspace{-12pt}
\caption{Distribution of the results for the E2 environment, according
to the mutation and default node action variables. See section~\ref{sub:signif} for
details and analysis.
 \label{fig:e2_results}}
\end{figure}

Box plots (see
figures~\ref{fig:maze10_results},~\ref{fig:e1_results}~and~\ref{fig:e2_results})
are provided to give an informal idea of the distributions of the
fitnesses for each environment, according to the two meaningful
variables (mutation and  default node action).
These graphics represent meaningful statistics rather than all the data:
the median as an horizontal line in the boxes, rather than the sample
mean, as the measure of central tendency due to the skew in the
distribution caused by the fixed lower limit of fitness values; the
second and third quartiles, taken as the upper and lower edges of the
boxes; finally, the ``whiskers'' are plot from both quartiles to the
extremal values.

The distribution for ATNoSFERES corresponds more or
less to the BitFlip/Random case. From these plots, one can see that, on the one hand,
Uniform mutation clearly outperforms BitFlip in all
cases, and on the other hand, for example for Maze10 (see
figure~\ref{fig:maze10_results}), that approximately 25\% of the 
Uniform/Random solutions outperform XCSMH (1st quartile). 

\subsection{Comparison of convergence time with ACS and XCSMH}\label{sub:compar_speed}

 Table~\ref{tab:comparison.old}, borrowed from
 \cite{Landau-Sigaud:IS:2004}, gives a comparison between the
 performance of the original \AoS\ and that of some well-known LCSs from the
 literature in terms of convergence time.

\begin{table*}
\begin{center}
\small
\begin{tabular}{|c|c||c|c|c|c|c||c|}\hline
envir. & LCS type & perf. & LCS & $PR$(\%) & $NG$ & $NT$ & NT/LCS\\
\small
\textbf{Maze10} & XCSM & 15.1 & 7,000 &100 & 8.42 & 45,000 & 6.42\\
\textbf{Maze10} & XCSMH & 6.1 & 6,500 & 7.30 & 4909 & 360.$10^6$ &55,000\\
\textbf{E1}  & ACS {\small ($BS_{max}$ = 1)} & 4 & 4,400  & 30.92 & 5115 & 218.$10^6$ & 50,000\\
\textbf{E2} & ACS {\small ($BS_{max}$ = 2 or 3)} & 6.5 & 2,000  &83.84 & 835 & 14.$10^6$ & 7,000\\
\hline
\end{tabular}
\caption{\footnotesize Comparison between the original \AoS\ and LCSs. The
  measure is the average number of trials needed to reach the
  performance given in column ``perf.''. 
$PR$ is the percentage of runs where \AoS\  outperforms the
corresponding LCS, and 
$NG$ is the average number of generations before this takes place.
Thus, with 300 individuals per generation, the average number of
evaluation runs 
necessary to outperform the corresponding LCS is $NE=300*NG*100/PR$,
and the average number of elementary runs $NT$ for an environment with
NS start 
cells ($NS=18$ for \textbf{Maze10}, $44$ for \textbf{E1} and $48$ for
\textbf{E2}) is: $NT=NS*NE$. 
Thus $NT$ gives the average number of elementary runs needed by \AoS\
to outperform the corresponding LCS. 
Finally, $NT/LCS$ gives a good approximation of the factor by which
the corresponding LCS is faster than \AoS\ to 
reach its best performance. 
Note that no performance comparison is given on \textbf{E1} against
ACS with $BS_{max} = 2$ since 
\AoS\ never outperforms it. 
}
\label{tab:comparison.old}
\end{center}
\end{table*}

From this table, it was clear that, though being more general and
obtaining sometimes better solutions, \AoS\ was several orders of
magnitude slower than all other tested LCSs.
However, since the modifications that lead to ATNoSFERES-II
significantly improved those performances, it remains to check
again whether this improvement has a significant impact on the running
time of the algorithm, and to compare this running time to that of the
same LCSs. The results are shown in table~\ref{tab:comparison.new},
and show that ATNoSFERES-II is indeed much faster than ATNoSFERES,
though still slower than the other LCSs. 

\begin{table*}
\begin{center}
\small
\begin{tabular}{|c|c||c|c|c|c|c||c|}\hline
envir. & LCS type & perf. & LCS & $PR$(\%) & $NG$ & $NT$ & NT/LCS\\
\hline
\textbf{Maze10} & XCSM & 15.1 & 7,000 & 100 & 10  & 54,000 & 7.7 \\
\textbf{Maze10} & XCSMH & 6.1 & 6,500 & 27.5 & 5,300 & 104.$10^6$ & 16,000 \\
\textbf{E1} & ACS {\small ($BS_{max}$ = 1)} & 4 & 4,400  & 92 & 2,500 & 36.$10^6$  & 8,100 \\
\textbf{E1} & ACS {\small ($BS_{max}$ = 2)} & 3.3 & 1,200  & 22.5 & 6,900 & 40.$10^7$ & 337,000 \\
\textbf{E2} & ACS {\small ($BS_{max}$ 2 or 3)} & 6.5 & 2,000  & 99 & 460 & 67.$10^5$  & 3,300 \\
\hline
\end{tabular}
\caption{\footnotesize Same comparison as in table~\ref{tab:comparison.old} between
  the new version of \AoS\ (assuming only Uniform mutation and Random default node action), and LCSs.
This time, performance comparisons can be given on \textbf{E1} against
ACS with $BS_{max} = 2$.}
\label{tab:comparison.new}
\end{center}
\end{table*}

\section{Discussion}\label{sec:discussion}

The first remark we want to make is that, despite the improvement in
performance brought by our modifications, we still did not succeed in
reaching the optimal policy in any of the three environments tested
here. This illustrates how difficult these simple non Markov
environments are for a 
genetic-based machine learning system.

From section~\ref{sub:signif}, we can see that the most important and
significant improvement is the change of representation, through the 
induced change of mutation. The impressive efficiency of this
modification is probably 
mainly due to the codons-to-tokens mapping that was used in the
original version of \AoS\  (table~\ref{tab:genetic-code}). 
As a matter of fact, one can note that
the leftmost bit is an indicator for label tokens: tokens enumerated
from 100000 to 111111 are exclusively action or condition tokens, to be
pushed on the stack. Therefore, when using a bit-flip mutation over the
binary encoding of the codons, with a 1\% probability, once a codon has its
leftmost bit set, it has 99\% probabilities to remain an 
action or a condition:  the combination of the
mapping and the bit-flip mutation resulted in a very strong bias
toward labels exploration, with very little structure
exploration.
On the other hand, when using the uniform mutation over string of
integers, structural changes occur with probability 0.5 during
mutation -- and the results 
show that this additional structural exploration
significantly improves the performance of the algorithm. 
Further work will look in detail at the statistics of every token
used by the best solutions, and should help to
find a more efficient distribution bias over the different tokens,
that will be easy to implement in the integer representation.

The introduction of
\emph{Finish} default action node did not prove very useful: either
\emph{Random} 
performs significantly better, or the difference is not
statistically significant.  
But this conclusion is consistent with the general consensus in the
Reinforcement Learning community, according to which some
non-determinism is always helpful when tackling Non Markov problems, as
purely 
deterministic controllers might easily get stuck in local minima.
However, in the context of evolutionary techniques, non-determinism makes
the evaluation more difficult and may be detrimental to the
convergence of the GA. Some optimal trade-off probably remains to be
found here.

The \emph{Node/Label} set
of stack tokens, introduced in order to facilitate structural
modifications of the automaton being built during the interpretation
process, only improves results in the Maze10 environment. 
Restricting to the Uniform/Random combination, this advantage vanishes
(see table~\ref{tab:p-values2}). More precisely, on Maze10, only restricting 
to \emph{Uniform} results in \emph{Random} still significantly
outperforming  {\em Finish} on average (see table~\ref{tab:p-values2}), and such is the
case too for \emph{Node/Label} 
with respect to \emph{All}, with even more significance. So, the
incompatibility on Maze10 lies between \emph{Node/Label}  and
\emph{Random}.

Nevertheless, this phenomenon is difficult to explain. In
\cite{Landau-Sigaud:IS:2004}, it was  shown that the kind of
structures necessary to perform optimally in Maze10, E1 and E2 are
very different from one another. It is likely that a difference in the
token language would result in a different probability of obtaining
this or that structure, but the corresponding analysis is at the
moment beyond our reach. If this assumption appears to be
true, this means that the token language itself has to be specifically
tuned for each particular problem.

Quite surprisingly, the \emph{Contradiction/Non-Contradiction}
alternative 
does not systematically make significant difference either. As a
matter of fact, the difference is significant in
average fitnesses for both alternatives only on E1. It could have
been expected that biasing the population towards meaningful condition
parts would  systematically reduce the search space towards
efficient automata, but this does not seem to be the case, even if 
the Non-Contradictory language seems to perform slightly
better in all experiments.
Here, we must take into account the fact that the Non-Contradictory language
induces more introns in the genotype, as many conditions are
simply ignored once put upon an edge. So again, there is probably a
trade-off between a 
more efficient search, thanks to more meaningful edges, and
degradation of performances 
because of the genotype bloat.  Future work will investigate 
this assumption by trying much shorter
genotypes, leaving less room to introns, and/or adding some parsimony
pressure to the fitness. If our assumption is true, we
should observe even better results on average when exploring a space
biased towards shorter genotypes.

Finally, from the results of section~\ref{sub:compar_speed}, it is
clear that, even if the improvement in performance has a serious
impact on convergence time, this impact is still far form sufficient
to make ATNoSFERES-II competitive in speed with ACS,
\cite{stolzmann:latent,metivier:lattaud:ACS}
and XCSMH \cite{lanzi00_toward},  the most 
efficient Michigan-style LCSs (the case of XCSM can be ignored due to
the much higher efficiency of XCSMH).
We feel that this conclusion advocates once again for the more general
claim that:
\begin{itemize}
\item the Pittsburgh approaches are often more robust than
Michigan ones and, given enough time, can often obtain better
performances, but
\item they are generally much slower, which results in poorer
performance under strong CPU time constraints.
\end{itemize}

\section{Conclusion}\label{sec:conclusion}
In this paper, we have studied the effect on the performance of
several modifications of the Pittsburgh style system \AoS.
Some of these modifications, such as a different encoding of the token
language, have been demonstrated to have a statistically significant
impact 
on the performance. Other modifications, such as the deterministic
versus 
stochastic behavior of the automaton, or the presence versus absence of
contradictions on the label of edges, have given less conclusive
results. 

In some cases, the effect of these variations have not been clearly
explained so far: further investigations are necessary to change
the assumptions we made about these phenomena into unquestionable
explanations. 

Anyway, one clear result of this paper is that, thanks to some of
these modifications, ATNoSFERES-II was able to find the best CS-based
controllers 
known so far in the Maze10, E1 and E2 environments. However, it should
be noticed that 
this improvement in performance was not accompanied by a clear
improvement in convergence speed.
This last assertion, as well as general considerations known as the
Michigan vs Pittsburgh debate, are a strong incentive to try to
design a Michigan-style system based on the same representation as
ATNoSFERES. We hope that the studies conducted here will help in 
making this future system more efficient.

\vfill\eject
\newcommand{\noopsort}[1]{} \newcommand{\printfirst}[2]{#1}
  \newcommand{\singleletter}[1]{#1} \newcommand{\switchargs}[2]{#2#1}
  \newcommand{\guileft}{``} \newcommand{\guiright}{''} \newcounter{saeclum}
  \newcommand{\siecle}[1]{\setcounter{saeclum}{#1}\textsc{\roman{saeclum}}\fup%
{e}}


\begin{thebibliography}{10}
\vspace*{0.5mm}
\scriptsize

\bibitem{landau:picault:SBGE:eng}
S.~Landau and S.~Picault.
\newblock Stack-{B}ased {G}ene {E}xpression.
\newblock Technical report {LIP}6 2002/011, {LIP}6, Paris, 2002.

\bibitem{landau:picault:atnosferes}
S.~Landau, S.~Picault, and A.~Drogoul.
\newblock {ATNoSFERES}: a {M}odel for {E}volutive {A}gent {B}ehaviors.
\newblock In {\em Proceedings of the {AISB}'01 Symposium on Adaptive Agents and
  Multi-Agent Systems}, 2001.

\bibitem{Landau:GECCO:2002}
S.~Landau, S.~Picault, O.~Sigaud, and P.~G{\'e}rard.
\newblock A comparison between {ATNoSFERES} and {XCSM}.
\newblock In {\em GECCO 2002: Proceedings of the Genetic and Evolutionary
  Computation Conference}, pages 926--933, New York, 9-13 July 2002. Morgan
  Kaufmann Publishers.

\bibitem{Landau-Sigaud:IS:2004}
S.~Landau and O.~Sigaud.
\newblock {A} {C}omparison between {ATNoSFERES} and {LCSs} on {non-Markov}
  problems \emph{(to appear)}.
\newblock {\em Information Sciences}, 2004.

\bibitem{landau:IWLCS:2003}
S.~Landau, O.~Sigaud, S.~Picault, and P.~G{\'e}rard.
\newblock {A}n {E}xperimental {C}omparison between {ATNoSFERES} and {ACS}.
\newblock In W.~Stolzmann et~al., editors, {\em {IWLCS-03}. {P}roceedings of
  the {S}ixth {I}nternational {W}orkshop on {L}earning {C}lassifier {S}ystems},
  LNAI, Chicago, july 2003. Springer.

\bibitem{lanzi98analysis}
P.-L. Lanzi.
\newblock An analysis of the memory mechanism of {XCSM}.
\newblock In J.~R. Koza, W.~Banzhaf, K.~Chellapilla, K.~Deb, M.~Dorigo, D.~B.
  Fogel, M.~H. Garzon, D.~E. Goldberg, H.~Iba, and R.~Riolo, editors, {\em
  Genetic Programming 1998: Proceedings of the Third Annual Conference}, pages
  643--651, University of Wisconsin, Madison, Wisconsin, USA, 22-25 1998.
  Morgan Kaufmann.

\bibitem{ECJ:SpecialIssue2003}
P.-L. Lanzi, W.~Stolzmann, and S.~W. Wilson, editors.
\newblock {\em Special Issue of {\em Evolutionary Computation}}, volume 11:3.
\newblock MIT Press, 2003.

\bibitem{lanzi00_toward}
P.-L. Lanzi and S.~W. Wilson.
\newblock Toward optimal classifier system performance in non-markov
  environments.
\newblock {\em Evolutionary Computation}, 8(4):393--418, 2000.

\bibitem{metivier:lattaud:ACS}
M.~M{\'e}tivier and C.~Lattaud.
\newblock Anticipatory {C}lassifier {S}ystem using {B}ehavioral {S}equences in
  {N}on-{M}arkov {E}nvironments.
\newblock In Stolzmann et~al. \cite{IWLCS:2002}, pages 143--163.

\bibitem{EA03Nicolau}
M.~Nicolau, A.~Auger, and C.~Ryan.
\newblock Functional dependency and degeneracy: detailed analysis of the
  {GA}u{GE} system.
\newblock In {\em Evolution Artificielle 03}, pages 15--26. LNCS 2936, Springer
  Verlag, 2003.

\bibitem{stolzmann:latent}
W.~Stolzmann.
\newblock Latent {L}earning in {K}hepera {R}obots with {A}nticipatory
  {C}lassifier {S}ystems.
\newblock In Wu \cite{GECCO:1999}, pages 290--297.

\bibitem{IWLCS:2002}
W.~Stolzmann, P.-L. Lanzi, and S.~W. Wilson, editors.
\newblock {\em Proceedings of the {I}nternational {W}orkshop on {L}earning
  {C}lassifier {S}ystems ({IWLCS'02})}, LNAI, Granada, september 2002.
  Springer-Verlag.

\bibitem{wilson:animat:1}
S.~W. Wilson.
\newblock {C}lassifier {S}ystems and the {A}nimat {P}roblem.
\newblock {\em Machine Learning}, 2(3):199--228, 1987.

\bibitem{wilson89}
S.~W. Wilson and D.~E. Goldberg.
\newblock A critical review of {C}lassifier {S}ystems.
\newblock In {\em Proceedings of the Third International Conference on Genetic
  Algorithms}, pages 244--255, Los Altos, California, 1989. Morgan Kaufmann.

\bibitem{woods:atn}
W.~A. Woods.
\newblock {T}ransition {N}etworks {G}rammars for {N}atural {L}anguage
  {A}nalysis.
\newblock {\em Communications of the Association for the Computational
  Machinery}, 13(10):591--606, 1970.

\bibitem{GECCO:1999}
A.~S. Wu, editor.
\newblock {\em Proceedings of the 1999 Genetic and Evolutionary Computation
  Conference ({GECCO'99})}, 1999.

\end{thebibliography}

\end{document}